\documentclass{article}

\usepackage[square,numbers]{natbib}
\usepackage[preprint]{neurips_2019}
\usepackage{xcolor}
\usepackage[utf8]{inputenc} 
\usepackage[T1]{fontenc}    
\usepackage{hyperref}       
\usepackage{url}            
\usepackage{booktabs}       
\usepackage{amsfonts}       
\usepackage{nicefrac}       
\usepackage{microtype}      
\usepackage{algorithm}
\usepackage{algorithmic}

\title{Adversarial Attacks Against Deep Learning Systems for ICD-9 Code Assignment}

\author{%
  Sharan Raja$^{1,3}$, \textnormal{and} Rudraksh Tuwani$^{2,3}$ \\
  $^1$Massachusetts Institute of Technology, Cambridge, MA \\
  $^2$Harvard University, Cambridge, MA \\ 
  $^3$\texttt{\{rsharan, rudraksh\}@mit.edu} \\
}

\begin{document}

\maketitle

\begin{abstract}
  Manual annotation of ICD-9 codes is a time consuming and error-prone process. Deep learning based systems tackling the problem of automated ICD-9 coding have achieved competitive performance. Given the increased proliferation of electronic medical records, such automated systems are expected to eventually replace human coders. In this work, we investigate how a simple typo-based adversarial attack strategy can impact the performance of state-of-the-art models for the task of predicting the top 50 most frequent ICD-9 codes from discharge summaries. Preliminary results indicate that a malicious adversary, using gradient information, can craft specific perturbations, that appear as regular human typos, for less than $3\%$ of words in the discharge summary to significantly affect the performance of the baseline model.

\end{abstract}

\section{Introduction}
The International Classification of Diseases (ICD) establishes a standardized fine-grained classification system for a broad range of diseases, disorders, injuries, symptoms, and other related health conditions \cite{who}. It is primarily intended for use by healthcare workers, policymakers, insurers and national health program managers. The United States incurs administrative costs in billions of dollars annually arising from a complex billing infrastructure \cite{digital_health}. Specifically, the ICD code assignment is typically a manual process, consuming on average between 25 to 43 minutes per patient depending on the ICD version \cite{perspectives_2014}. It is also prone to errors resulting from inexperienced coders, variation between coders, incorrect grouping of codes or mistakes in the patient discharge summaries. These errors are very costly with one report estimating that preventable errors in ICD coding have cost Medicare system 31.6 billion in FY2018 \cite{cmi}.\\\\
Recent work \cite{mullenbach-etal-2018-explainable, Schmaltz2020ExemplarAF, amin} has tried to automate the task of ICD code assignment using deep learning.  Typically framed as a multilabel classification problem, researchers have trained Convolutional Neural Networks (CNN), Recurrent Neural Networks (RNN), and Transformer models to predict ICD-9 codes from patient discharge summaries.  These models have outperformed rule-based approaches and those utilizing conventional algorithms such as Logistic Regression, Support Vector Machines, Random Forests etc., achieving competitive micro F1-scores in the range 42\% - 68\%. Amongst these models, those based on CNNs have achieved the best performance. 

Neural network models have revolutionized the field of NLP and SOTA models for various NLP tasks involve deep neural network models such as BERT, Bidirectional RNN or CNN-based methods. Recent works \cite{kurakin2016adversarial,kurakin2016adversarialb,papernot2016practical,zhao2017generating} have shown a particular vulnerability of such deep models to adversarial examples that are often produced by adding small and imperceptible perturbations to the input data. The state of the art models of NLP are no exceptions to such perturbations. \cite{zhang2019adversarial} provides a review of different adversarial attacks and defense strategies in the NLP literature. Based on granularity of the perturbation, adversarial attack strategies in NLP can be classified into three types - character-level attacks, word-level attacks and sentence-level attacks. In a character-level attack strategy, the model induces noise at the character level. Character-level noise can be induced due to naturally occurring reasons such as typos and misspellings or due to intentional modification by a malicious third-party. \cite{Li_2019, eger2019text, ebrahimi2018adversarial} are some of the existing character-level attack strategies in NLP. To accurately model the naturally occurring typos, \cite{sun2020advbert} restrict the typos distribution based on the character constraints found in a standard English keyboard. We follow this strategy in our work. Furthermore, we assume a white-box setting where the adversary has access to gradients of the loss function wrt to the model inputs. To our knowledge, this is the first work to investigate the effects of adversarial samples in clinical NLP domain.  

\section{Data and Preprocessing}
We used MIMIC-III \cite{mimiciii}, a large open source database comprising information of patients admitted to critical care units of Beth Israel Deaconess Medical Center (Boston, Massachusetts, USA). The database contains de-identified Electronic health records with both structured and unstructured data including diagnostics and laboratory results, medications, and discharge summaries. In this work, we focus on discharge summaries which encapsulates details pertaining to a patient’s stay. \\\\
Each discharge summary is manually annotated by human coders with multiple ICD-9 codes, describing both the diagnoses and procedures that the patient underwent. Out of the approx. 13000 possible ICD-9 codes, 8921 (6918 diagnosis, 2003 procedure) are present in our dataset. Following previous work, we merge discharge summaries corresponding to the same patient ID, such that no patient appears twice in our dataset resulting in 47,427 discharge summaries. This is done to ensure that there is no ‘data leakage’ between train, validation, and test sets. \\\\
The full label setting is quite noisy and suffers from class imbalance. Potential sources of noise include both missed assignments (not annotating all relevant ICD-9 codes) and incorrect assignments (annotating similar but incorrect ICD-9 codes). Consequently, it is relatively trivial to develop an adversarial attack strategy in the full label setting. For instance, one could simply find the keywords corresponding to low frequency labels and then either append or remove them from a discharge summary to alter a machine learning model’s prediction. This strategy will however fail for frequent labels since we expect the model to generalize beyond simply memorizing a few keywords. Therefore, we limited the label set to the 50 most frequent labels and removed discharge summaries which were not annotated with at least one of the labels. The resulting dataset was then split into training, validation and testing sets which contained 8067, 1574, and 1730 discharge summaries, respectively.\\\\
We followed the same pre-processing steps as in previous work \cite{mullenbach-etal-2018-explainable}.  All tokens without any alphabetic characters were removed. We then lowercased all tokens and replaced those appearing less than three times in the training documents with an ‘UNK’ token. 

\label{headings}
\section{Baseline model}
Our baseline models were the same as \cite{mullenbach-etal-2018-explainable}.  Specifically, we used a CNN-based sentence classifier model introduced by \cite{kim} which utilizes a max pooling layer to get sentence vector representations. We call this model Max Pool based CNN. The other model that we use instead utilizes label embeddings to calculate attention weights over word positions. These weights are then used to pool the output of the convolutional layer and calculate the sentence vector representation. This model is referred to as the Attention Pool based CNN.

\label{others}
\section{Adversarial attack strategy}
\label{adv}

We generate adversarial examples based on the following algorithm: Given a pre-trained NLP algorithm $f : X \to y$ and a measure of classification $q: y \to s$, we are interested in finding perturbations $\delta x$ on the input $X$ such that $q(X + \delta x) \leq q(X)$ under the constraint $||\delta x || \leq K$. The final constraint ensures that the perturbations are small. In our work, we consider perturbations (typos) of four types:

\begin{enumerate}
    \item \textbf{Insert} - Insert characters into a word, such as hike $\to$ hlike
    \item \textbf{Delete} - Delete characters in a word, such as hike $\to$ hke
    \item \textbf{Swap} - swap two characters of a wors, such as hike $\to$ hkie
    \item \textbf{Replace} - Replace a character in a word with any neighboring keys in the keyboard, such as hike $\to$ hoke. Here o is a neighboring word to i in a standard english keyboard.
\end{enumerate}

Given an input sentence $s$ that is tokenized according to the model's tokenizer as $s = (w_1,w_2,..,w_N)$, we compute the partial derivative of loss with respect to each input item as shown below,  
\begin{equation}
    \mathcal{G}_{f}\left(w_{i}\right)=\nabla_{w_{i}} \mathcal{L}\left(w_{i}, y\right)
    \label{eq1}
\end{equation}

Based on this gradient information, we select a input word $w_i$ to attack. We experiment with two different strategies here, the maximum gradient strategy where we choose the word corresponding to the maximum gradient and a random strategy where a random word is chosen to attack. Once a word is chosen, we generate all possible typos based on the four ways described above. The typo which decreases the score of the output $y$ based on the score function $q$ is chosen. Here, we use the top5 precision as the score function. Now the word replaced with the optimal typo word is again fed through this loop for $K$ times. Each time, a different word is chosen to ensure that final words don't change from the initial words by a lot. We experiment with different choices of $K$. The algorithm is shown in alg.~\ref{alg:adv-ICD9}

\begin{algorithm}[t] 
\small
\caption{Adversarial attack for ICD-9 classification}
\label{alg:adv-ICD9}
\begin{algorithmic}[1]
\STATE{\textbf{Input:} Document $X$, ground truth labels $y*$, classifier $f(.)$, budget $K$ and score function $q$}
\STATE{$i \leftarrow 0, X_{best} \leftarrow X$}
\WHILE{$i \leq K$}
\STATE{$c \leftarrow$ Segmentation$(X_{best})$}
\FOR{each token $c_i$ in $c$}
\STATE{Compute gradients of component $c_i$ according to eq.~\ref{eq1}}
\ENDFOR
\STATE{Find the token or word based on the gradient according to the strategy}
\STATE{Generate all possible typos for the chosen word}
\STATE{Create a list of documents; each document corresponding to a typo}
\STATE{Find the document instance that decreases the output score the most assign this to $X_{best}$}
\STATE{$i \leftarrow i+1$}
\ENDWHILE
\end{algorithmic}
\end{algorithm}

\section{Results}

To the best of our knowledge, \cite{mullenbach-etal-2018-explainable} is the current state-of-the-art for the task of automated ICD-9 code assignment. We re-implemented their best performing models using the AllenNLP framework \cite{allennlp}. The test-set performance of the models for the task of predicting the top-50 most frequent ICD-9 codes from discharge summaries is given in table ~\ref{model-performance}. We found that the Max Pool based CNN outperformed the Attention Pool based CNN on all performance metrics. Further, we found that the computation time for training as well as generating predictions for the former was much lesser than the latter. Therefore, we decided to limit our focus on developing an adversarial attack strategy for the Max Pool based CNN. 

We experiment with three different values of budget $K = \{10, 20, 30\}$ and two different strategies - maximum gradient and random strategy for selecting the token to attack. The maximum gradient strategy can be used to analyze the robustness of the model to malicious attacks while the random strategy can be used to simulate natural settings with adversarial examples. The training time for each run on the entire corpus ($1725$ discharge summaries) - $8$ hrs to $16$ hrs on a machine with Tesla K80 GPU. The results are summarized in table ~\ref{adv-results}.

In accordance with our intuition, max grad strategy performs better than random strategy. This is because, max grad strategy can produce meaningful perturbations in a large input space (average size of input document is $\sim 1400$ tokens). The model's performance doesn't drop much with random strategy. This suggests that the model is some what robust to naturally occurring noise such as typos and misspellings. However, this might change as the budget is increased. Due to computational limits, we did not explore budgets beyond $30$. A key  result of our work is that, with less than $3\%$ of input tokens modified, the model's performance drops significantly from $0.62$ to $0.377$. This shows the potential vulnerability of this model to malicious attacks. Since, only a very few tokens are changed, it might be hard to defend against these attacks by training a discriminator to distinguish maliciously modified documents from regular ones.

Tables ~\ref{table1} and ~\ref{table2} show examples of discharge summaries before and after attack with their top5 labels. It is important to note that, on a few discharge summaries (last example in both the tables), the algorithm increases the top5 precision instead of decreasing it. One can make modifications to the algorithm to ensure that this doesn't happen which would result in further drop in precision. Due to time constraints, we were not able to accommodate this modification. Nevertheless, these examples show the brittleness of the baseline model to input tokens.  

\begin{table}
  \caption{Performance of baseline models on MIMIC-III dataset for predicting the top 50 most frequent ICD-9 codes.}
  \label{model-performance}
  \centering
  \begin{tabular}{lcc}
    \toprule
    \multicolumn{3}{c}{Model}                   \\
    \cmidrule(r){1-3}
    Metric     & Max Pool CNN & Label Attention Pool CNN \\
    \midrule
    \cmidrule(r){1-3}
    Macro F1 Score     & $0.55$ & $0.49$  \\
    Micro F1 Score     & $0.63$ & $0.55$  \\
    Macro AUC     & $0.87$ & $0.83$  \\
    Micro AUC    & $0.91$ & $0.86$  \\
    Top 5 Precision     & $0.62$ & $0.54$  \\
    \bottomrule
  \end{tabular}
\end{table}

\begin{table}
  \caption{Results of adversarial attacks on the corpus of discharge summaries of size $1725$.}
  \label{adv-results}
  \centering
  \begin{tabular}{ccc}
    \toprule
    \multicolumn{3}{c}{Top5 precision}                   \\
    \cmidrule(r){1-3}
    Budget     & Max grad strategy & Random strategy \\
    \midrule
    \multicolumn{3}{c}{Baseline ($K = 0 $) $\to 0.62$}\\
    \cmidrule(r){1-3}
    $10$     & $0.549$ & $0.592$  \\
    $20$     & $0.462$ & $0.574$  \\
    $30$     & $0.377$ & $0.567$  \\
    \bottomrule
  \end{tabular}
\end{table}

\begin{table}
  \caption{Examples of sentences for budget $10$ in maximum gradient strategy where the adversarial attack strategy resulted in maximum change in top5 labels. The first two examples cause the predictions to be worse and the last example shows a case where the adversarial example results in increased top5 precision. Labels in blue appear are part of ground truth labels.}
  \label{table1}
  \centering
  \begin{tabular}{ll}
    \toprule
    \multicolumn{2}{c}{Maximum gradient strategy, budget $= 10$}                   \\
    \cmidrule(r){1-2}
    Top5 precision & Description  \\
    \midrule
    $0.8 \to 0.2$ & \begin{tabular}{@{}l@{}}...unchanged as well. A tracheostomy tube and right subclavian line... \\ ...unchanged as well. A \textcolor{red}{\textbf{tacheostomy ttube}} and right subclavian line...\\ \\ ...performed on. During tracheostomy procedure, pneumothorax occured and \\ chest tube...\\...performed on. During \textcolor{red}{\textbf{tacheostomy} \textbf{proecedure}, \textbf{pneumothroax} \textbf{occurred}} \\ and chest tube...\\ \\\textbf{Top5 labels before attack} - \color{blue} Insertion of Sengstaken tube, \color{red} Pneumonia, \\ \color{blue}  Respiratory Ventilation, Venous catheterization, Arterial catheterization \\\\ \textbf{Top5 labels after attack} - \color{red} Pneumonia, Unspecified pleural effusion, \color{blue} Insertion \\ \color{blue} of Sengstaken tube, \color{red} Anemia, Acute post-hemorrhagic anemia\end{tabular}\\
    \cmidrule(r){1-2}
    $0.8 \to 0.2$ & \begin{tabular}{@{}l@{}}...cholelithiasis complicated hospital course including sepsis w persistent \\ hyperbilirubinemia... \\ ...cholelithiasis complicated hospital course including \textcolor{red}{\textbf{sespis}} w persistent \\ hyperbilirubinemia...\\ \\ ...surgical or invasive procedure - ercp, laparoscopic cholecystectomy, \\ laparoscopic liver biopsy..\\...surgical or invasive \textcolor{red}{\textbf{preocedure}} - \textcolor{red}{\textbf{erccp, laproscopic, cholecysectomy}},\\ laparoscopic liver biopsy..\\\\ ...presentation to hospital1 intubated jaundiced scleral...\\...presentation to hospital1 \textcolor{red}{\textbf{int8bated}} jaundiced scleral...\\ \\\textbf{Top5 labels before attack} - \color{blue} Unspecified acquired hypothyroidism, Insertion \\ \color{blue} of endotracheal tube, Respiratory Ventilation, Enteral infusion of concentrated \\  \color{blue} nutritional substances, \color{red} Continuous invasive mechanical ventilation \\\\\textbf{Top5 labels after attack} - \color{blue} Unspecified acquired hypothyroidism, \color{red} Diagnostic \\ \color{red} ultrasound of heart, Old myocardial infarction, Major depressive disorder,\\ \color{red} Other and unspecified hyperlipidemia.\end{tabular}\\
    \cmidrule(r){1-2}
    \textcolor{red}{\textbf{$0.2 \to 0.8$}} & \begin{tabular}{@{}l@{}}...higher on tube feeds appreciate nutrition recs tfs changed to...\\ ...higher on \textcolor{red}{\textbf{ttube fees apprciate nutritin res tfts}} changed to..\\ \\ ...for both chf and suspected aspiration pna w iv lasix...\\...for both chf and suspected \textcolor{red}{\textbf{aspirtation}} pna w iv lasix...\\\\ \textbf{Top5 labels before attack} - \color{red} Enteral infusion of concentrated nutritional \\ \color{red} substances \color{blue} Venous catheterization, \color{red} Food / vomit pneumonitis, Urinary tract \\ \color{red} infection, Acute respiratory failure.\\\\ \textbf{Top5 labels after attack} - \color{red} Acute respiratory failure, \color{blue} Venous catheterization, \\ \color{blue} Congestive heart failure, \color{blue} Insertion of endotracheal tube, Unspecified essential \\ \color{blue} hypertension \end{tabular}\\
    \bottomrule
  \end{tabular}
\end{table}

\begin{table}
  \caption{Examples of sentences for budget $20$ in maximum gradient strategy where the adversarial attack strategy resulted in maximum change in top5 labels. The first two examples cause the predictions to be worse and the last example shows a case where the adversarial example results in increased top5 precision. Labels in blue appear are part of ground truth labels.}
  \label{table2}
  \centering
  \begin{tabular}{ll}
    \toprule
    \multicolumn{2}{c}{Maximum gradient strategy, budget $= 20$}                   \\
    \cmidrule(r){1-2}
    Top5 precision & Description  \\
    \midrule
    $1.0 \to 0.2$ & \begin{tabular}{@{}l@{}}...cabg, x4, hyperlipidemia, anxiety, hypertension, migraines, gi bleed... \\ ...\textcolor{red}{\textbf{cbg}}, x4, \textcolor{red}{\textbf{hyperlipiddemia, axiety, hypertnesion, migranes}}, gi, \textcolor{red}{\textbf{bleedd}}...\\ \\ ...medical history - coronary artery disease, hyperlipidemia, anxiety...\\...medical history - \textcolor{red}{\textbf{conronary atery}} disease, \textcolor{red}{\textbf{hyperlipdiemia}}, anxiety...\\\\ ...room and underwent coronary artery bypass grafting x4 with left...\\...room and underwent coronary \textcolor{red}{\textbf{bypas gratfting}} x4 with left...\\ \\\textbf{Top5 labels before attack} - \color{blue} Single internal mammary-coronary artery bypass,  \\ \color{blue} Extracorporeal circulation auxiliary to open heart surgery, Other and unspecified\\ \color{blue} hyperlipidemia, Atherosclerotic heart disease of native coronary artery without \\ \color{blue} angina pectoris, Unspecified essential hypertension \\\\ \textbf{Top5 labels after attack} - \color{blue} Extracorporeal circulation auxiliary to open heart\\ \color{blue} surgery, \color{red} Enteral infusion \color{red} of concentrated nutritional substances, Transfusion of \\ \color{red} packed cells, Diagnostic ultrasound of heart, \color{red} Atrial fibrillation\end{tabular}\\
    \cmidrule(r){1-2}
    $1.0\to 0.2$ & \begin{tabular}{@{}l@{}}...to posterior descending artery bronchosccopy reintubated history of present...\\ ...to posterior descending artery bronchosccopy \textcolor{red}{\textbf{reitnubated}} history of present...\\ \\ ...the procedure was hemoptysis requiring intubation he was transferred back...\\...the procedure was hemoptysis requiring \textcolor{red}{\textbf{ibntubation}} he was transferred back...\\\\ ...mitral regurgitation, hypertension, hypercholesterolemia, congestive heart\\ failure, tobacco abuse...\\...\textcolor{red}{\textbf{motral regunrgitation,}} hypertension, hypercholesterolemia, \textcolor{red}{\textbf{congesitve heeart}}\\\textcolor{red}{\textbf{failre, taobacco abusee}} \\\\\\\textbf{Top5 labels before attack} - \color{blue} Extracorporeal circulation auxiliary to open heart\\ \color{blue} surgery,  Single internal mammary-coronary artery bypass, Atherosclerotic heart\\ \color{blue} disease of native coronary artery without angina pectoris, Mitral valve disorders,\\ \color{blue} Congestive heart failure \\\\\textbf{Top5 labels after attack} - \color{red} Unspecified essential hypertension, enteral infusion of  \\ \color{red} concentrated nutritional substances, \color{blue} Extracorporeal circulation auxiliary to open\\ \color{blue} heart surgery, \color{red} Respiratory Ventilation, Transfusion of packed cells.\end{tabular}\\
    \cmidrule(r){1-2}
    \textcolor{red}{\textbf{$0.4 \to 1.0$}} & \begin{tabular}{@{}l@{}}...cancer s p resection bilateral renal masses per pcp name...\\ ...cancer s p \textcolor{red}{\textbf{resecton bliateral}} \textcolor{red}{\textbf{reanl mases}} per pcp name..\\ \\ ...morbid obesity, depression, restless leg syndrome...\\...\textcolor{red}{\textbf{mtorbid obestity, deprssion, resltess}} leg syndrome...\\\\ \textbf{Top5 labels before attack} - \color{blue} Congestive heart failure, Chronic obstructive \\ \color{blue} pulmonary disease \color{red} Chronic kidney disease, Hypertensive chronic kidney disease,\\ \color{red} Non-invasive mechanical ventilation\\\\ \textbf{Top5 labels after attack} - \color{blue} Congestive heart failure, Chronic obstructive pulmonary \\ \color{blue} disease, Unspecified essential hypertension, Diabetes mellitus without mention of \\ \color{blue} complication, Urinary tract infection \end{tabular}\\
    \bottomrule
  \end{tabular}
\end{table}

\section{Discussion}
This work is a first step at exploring the robustness of NLP models used for automatic ICD-9 code classification. Clinical documents are different from regular documents as they are typically generated in a fast-paced environment with higher than average typos and non-standard acronyms. As a result, clinical NLP models are more susceptible to adversarial samples compared to a regular NLP model trained on a standard English dataset. A key extension of the work would be to consider a dictionary learnt from clinical documents and biomedical literature as a defense against these character-level perturbations. Although this might mitigate the decrease in performance, it wouldn't completely solve it. A more rigorous way to deal with this would be to account for this in the tokenization strategy. It is easy to push a word out of vocabulary when using tokenization strategies like word2vec and GloVe. Other strategies that model words unseen in training dataset such as word-piece and byte-pair encoding will also break when typos are introduced because these models learn sub words from a standard dictionary. Therefore, any defense must account for these typos in the fundamental tokenization strategy. An interesting direction would be to learn a word similarity metric and map an unknown word to a closer word in the vocabulary given the input word and the context in which it appears. Building a robust tokenization strategy would be the first step towards a robust NLP model against character-level adversarial attacks.   

\medskip

\small

\bibliography{main}

\begin{thebibliography}{19}
\providecommand{\natexlab}[1]{#1}
\providecommand{\url}[1]{\texttt{#1}}
\expandafter\ifx\csname urlstyle\endcsname\relax
  \providecommand{\doi}[1]{doi: #1}\else
  \providecommand{\doi}{doi: \begingroup \urlstyle{rm}\Url}\fi

\bibitem[per(2014)]{perspectives_2014}
Perspectives, 2014.
\newblock URL
  \url{https://perspectives.ahima.org/preparing-for-icd-10-cmpcs-implementation-impact-on-productivity-and-quality/}.

\bibitem[cmi(2018)]{cmi}
Error rate drops, but medicare still lost \$31.6 billion to preventable billing
  errors in fy2018, 2018.

\bibitem[who(2019)]{who}
International classification of diseases (icd) information sheet, Oct 2019.
\newblock URL \url{https://www.who.int/classifications/icd/factsheet/en/}.

\bibitem[Amin et~al.(2019)Amin, Neumann, Dunfield, Vechkaeva, Chapman, and
  Wixted]{amin}
S.~Amin, G.~Neumann, K.~Dunfield, A.~Vechkaeva, K.~Chapman, and M.~Wixted.
\newblock Mlt-dfki at clef ehealth 2019: Multi-label classification of icd-10
  codes with bert.
\newblock 09 2019.

\bibitem[Ebrahimi et~al.(2018)Ebrahimi, Lowd, and Dou]{ebrahimi2018adversarial}
J.~Ebrahimi, D.~Lowd, and D.~Dou.
\newblock On adversarial examples for character-level neural machine
  translation, 2018.

\bibitem[Eger et~al.(2019)Eger, Şahin, Rücklé, Lee, Schulz, Mesgar,
  Swarnkar, Simpson, and Gurevych]{eger2019text}
S.~Eger, G.~G. Şahin, A.~Rücklé, J.-U. Lee, C.~Schulz, M.~Mesgar,
  K.~Swarnkar, E.~Simpson, and I.~Gurevych.
\newblock Text processing like humans do: Visually attacking and shielding nlp
  systems, 2019.

\bibitem[Gardner et~al.(2018)Gardner, Grus, Neumann, Tafjord, Dasigi, Liu,
  Peters, Schmitz, and Zettlemoyer]{allennlp}
M.~Gardner, J.~Grus, M.~Neumann, O.~Tafjord, P.~Dasigi, N.~Liu, M.~Peters,
  M.~Schmitz, and L.~Zettlemoyer.
\newblock Allennlp: A deep semantic natural language processing platform.
\newblock 03 2018.

\bibitem[Johnson et~al.(2016)Johnson, Pollard, Shen, Li-wei, Feng, Ghassemi,
  Moody, Szolovits, Celi, and Mark]{mimiciii}
A.~E. Johnson, T.~J. Pollard, L.~Shen, H.~L. Li-wei, M.~Feng, M.~Ghassemi,
  B.~Moody, P.~Szolovits, L.~A. Celi, and R.~G. Mark.
\newblock Mimic-iii, a freely accessible critical care database.
\newblock \emph{Scientific data}, 3:\penalty0 160035, 2016.

\bibitem[Kim(2014)]{kim}
Y.~Kim.
\newblock Convolutional neural networks for sentence classification.
\newblock \emph{Proceedings of the 2014 Conference on Empirical Methods in
  Natural Language Processing}, 08 2014.
\newblock \doi{10.3115/v1/D14-1181}.

\bibitem[Kurakin et~al.(2016{\natexlab{a}})Kurakin, Goodfellow, and
  Bengio]{kurakin2016adversarial}
A.~Kurakin, I.~Goodfellow, and S.~Bengio.
\newblock Adversarial examples in the physical world, 2016{\natexlab{a}}.

\bibitem[Kurakin et~al.(2016{\natexlab{b}})Kurakin, Goodfellow, and
  Bengio]{kurakin2016adversarialb}
A.~Kurakin, I.~Goodfellow, and S.~Bengio.
\newblock Adversarial machine learning at scale, 2016{\natexlab{b}}.

\bibitem[Li et~al.(2019)Li, Ji, Du, Li, and Wang]{Li_2019}
J.~Li, S.~Ji, T.~Du, B.~Li, and T.~Wang.
\newblock Textbugger: Generating adversarial text against real-world
  applications.
\newblock \emph{Proceedings 2019 Network and Distributed System Security
  Symposium}, 2019.
\newblock \doi{10.14722/ndss.2019.23138}.
\newblock URL \url{http://dx.doi.org/10.14722/ndss.2019.23138}.

\bibitem[Mullenbach et~al.(2018)Mullenbach, Wiegreffe, Duke, Sun, and
  Eisenstein]{mullenbach-etal-2018-explainable}
J.~Mullenbach, S.~Wiegreffe, J.~Duke, J.~Sun, and J.~Eisenstein.
\newblock Explainable prediction of medical codes from clinical text.
\newblock In \emph{Proceedings of the 2018 Conference of the North {A}merican
  Chapter of the Association for Computational Linguistics: Human Language
  Technologies, Volume 1 (Long Papers)}, pages 1101--1111, New Orleans,
  Louisiana, June 2018. Association for Computational Linguistics.
\newblock \doi{10.18653/v1/N18-1100}.
\newblock URL \url{https://www.aclweb.org/anthology/N18-1100}.

\bibitem[Papernot et~al.(2016)Papernot, McDaniel, Goodfellow, Jha, Celik, and
  Swami]{papernot2016practical}
N.~Papernot, P.~McDaniel, I.~Goodfellow, S.~Jha, Z.~B. Celik, and A.~Swami.
\newblock Practical black-box attacks against machine learning, 2016.

\bibitem[Schmaltz and Beam(2020)]{Schmaltz2020ExemplarAF}
A.~Schmaltz and A.~L. Beam.
\newblock Exemplar auditing for multi-label biomedical text classification.
\newblock \emph{ArXiv}, abs/2004.03093, 2020.

\bibitem[Shull(2018)]{digital_health}
J.~Shull.
\newblock Digital health and the state of interoperable ehrs (preprint).
\newblock \emph{JMIR Medical Informatics}, 7, 11 2018.
\newblock \doi{10.2196/12712}.

\bibitem[Sun et~al.(2020)Sun, Hashimoto, Yin, Asai, Li, Yu, and
  Xiong]{sun2020advbert}
L.~Sun, K.~Hashimoto, W.~Yin, A.~Asai, J.~Li, P.~Yu, and C.~Xiong.
\newblock Adv-bert: Bert is not robust on misspellings! generating nature
  adversarial samples on bert, 2020.

\bibitem[Zhang et~al.(2019)Zhang, Sheng, Alhazmi, and Li]{zhang2019adversarial}
W.~E. Zhang, Q.~Z. Sheng, A.~Alhazmi, and C.~Li.
\newblock Adversarial attacks on deep learning models in natural language
  processing: A survey, 2019.

\bibitem[Zhao et~al.(2017)Zhao, Dua, and Singh]{zhao2017generating}
Z.~Zhao, D.~Dua, and S.~Singh.
\newblock Generating natural adversarial examples, 2017.

\end{thebibliography}
\end{document}